\documentclass[lettersize,journal]{IEEEtran}
\usepackage{amsmath,amsfonts}
\usepackage{algorithm}
\usepackage{array}
\usepackage[caption=false,font=normalsize,labelfont=sf,textfont=sf]{subfig}
\usepackage{textcomp}
\usepackage{stfloats}
\usepackage{url}
\usepackage{verbatim}
\usepackage{graphicx}
\usepackage{cite}

\usepackage{array}
\usepackage{graphicx}
\usepackage{multirow}
\usepackage{multicol}
\usepackage{algpseudocode}
\usepackage{algorithmicx}
\usepackage{afterpage}
\usepackage{booktabs}

\usepackage[most]{tcolorbox}
\usepackage{afterpage}
\usepackage{tabularx}
\usepackage{calc}
\begin{document}

\title{Empathy Level Prediction in Multi-Modal Scenario with Supervisory Documentation Assistance}

\author{Yufei Xiao and Shangfei Wang,~\IEEEmembership{Member,~IEEE,}}

\markboth{IEEE TRANSACTIONS ON AFFECTIVE COMPUTING}%
{Shell \MakeLowercase{\textit{et al.}}: A Sample Article Using IEEEtran.cls for IEEE Journals}

\maketitle

\begin{abstract}
Prevalent empathy prediction techniques primarily
concentrate on a singular modality, typically textual, thus neglecting multi-modal processing capabilities. 
They also overlook the utilization of certain privileged information, which may encompass additional 
empathetic content. In response, we introduce an advanced multi-modal empathy prediction method integrating
video, audio, and text information. The method comprises the Multi-Modal Empathy Prediction and Supervisory 
Documentation Assisted Training. We use pre-trained networks in the empathy prediction network to extract 
features from various modalities, followed by a cross-modal fusion. This process
yields a multi-modal feature representation, which is employed
to predict empathy labels. To enhance the extraction of
text features, we incorporate supervisory documents as privileged
information during the assisted training phase. Specifically,
we apply the Latent Dirichlet Allocation model to identify potential
topic distributions to constrain text features. These supervisory
documents, created by supervisors, focus on the counseling topics
and the counselor’s display of empathy. Notably, this privileged
information is only available during training and is not accessible
during the prediction phase. Experimental results on the multi-modal and dialogue empathy datasets 
demonstrate that our approach is superior to the existing methods.
\end{abstract}


\begin{figure*}[!ht]
\centering
\includegraphics[width=\linewidth]{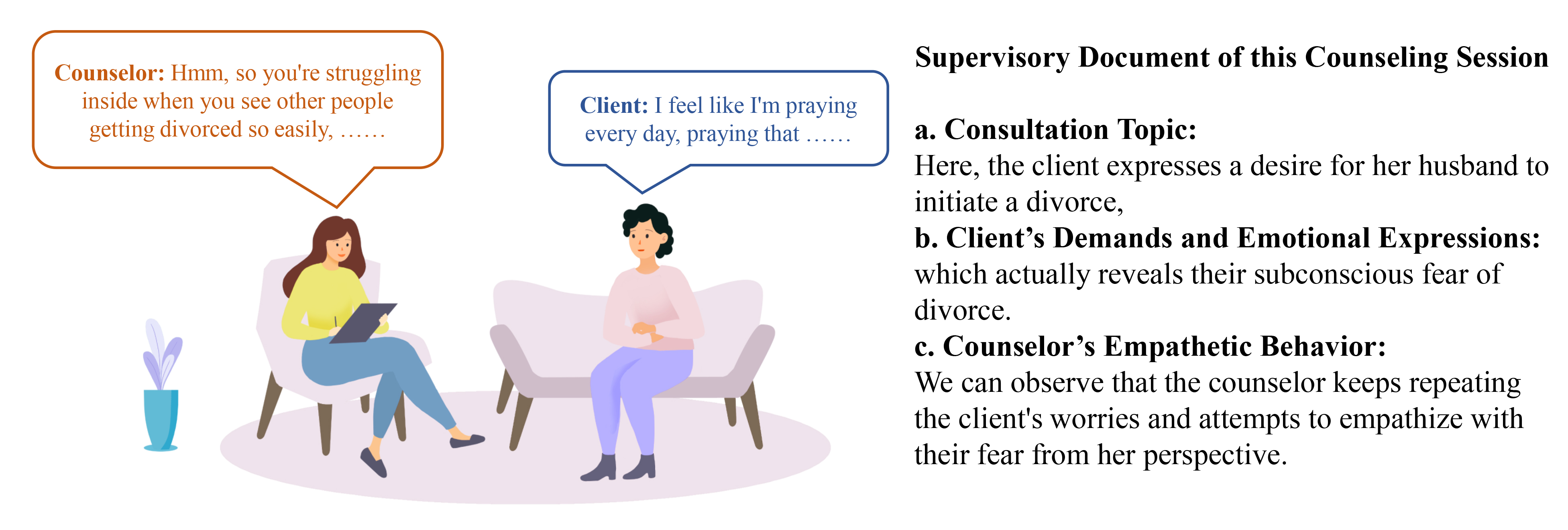}
\caption{An example of the supervisory documents. The left is the conversation in the counseling 
session, and the right is the corresponding supervisory document}
\label{fig:label_example}
\end{figure*}

\section{Introduction}
\IEEEPARstart{E}{mpathy} is characterized by an emotional response that arises from the 
interplay between inherent 
traits and situational factors. This empathetic response is spontaneously triggered, yet it is also 
sculpted by deliberate cognitive control. The emergent emotion closely mirrors the individual's direct 
or imagined experience and cognitive understanding of the emotional stimulus, accompanied by an awareness 
that the origin of this emotion lies outside oneself~\cite{cuff2016empathy}.
It is an essential element of human communication and interaction that enables our profound 
understanding and connection with others. It requires the capacity to discern and respond to 
the sentiments and perspectives of others. Therefore, predicting empathy plays a significant 
role in a wide range of applications, including, but not limited to mental health support, social media, and 
human-robot interaction. For instance, for mental health care, precise empathy prediction can assist 
therapists in adapting their approaches, aligning more closely with the emotional needs of clients, 
resulting in more effective and personalized treatments. 


Previous studies on predicting empathy aimed at designing efficient network 
architectures for predicting empathy levels
\cite{zhou2021language,montiel2022explainable,lahnala2022caisa,vettigli2021empna,shi2021modeling}.
They modeled the interaction between the client and the counselor during counseling 
sessions into feature representations to predict empathy ratings~\cite{wu2021towards,wu2022towards,lee2022formality}.  
Although previous studies have made significant progress in empathy prediction, they mainly 
focused on using unimodal input, particularly text. Most real-world
scenarios requiring empathy understanding, such as mental counseling, usually involve multi-modal input, 
including text, audio, and video. The overreliance on singular modalities leads to the incomprehensive 
understanding of empathy. 

Furthermore, few works in empathy prediction involve the utilization of privileged 
information, which may contain additional empathetic content. For example, the supervisory 
documents are objective descriptions of the entire counseling process conducted by a supervisor, 
which is independent of mental counseling. As illustrated in Figure 1, the descriptions mainly 
include topics and backgrounds of the counseling session, such as family, work, marriage, and others, 
as well as the client's demands and emotional expressions, and the counselor's empathetic behavior. 
Apart from these objective contexts and settings, this description typically includes the supervisor's 
subjective evaluation of the effectiveness of the counselor's empathetic behavior. Given their extensive 
expertise, psychological supervisors can offer clearer and more precise analyses of empathy in the 
counseling process. Therefore, the information contained in the supervisory documents, both subjective and 
objective, facilitates a more comprehensive analysis of empathetic expressions in counseling conversations, 
but they are exclusively accessible during the training phase, lacking presence in the testing phase and 
predictive downstream tasks.

\IEEEpubidadjcol

To address these limitations, we propose an advanced multi-modal empathy prediction method 
integrating video, audio, and text information. The method comprises two parts: Multi-Modal Empathy Prediction 
Network and Supervisory Documentation Assisted Training.  
The proposed empathy prediction network takes inputs from text, audio, and visual sources and generates a 
multi-modal representation to predict the empathy level. In addition, we utilize the supervisory documents to 
assist in the extraction of text features of the counseling process. We propose the supervisory 
documentation assisted training framework including a Latent Dirichlet Allocation (LDA) module to extract 
the topic distribution of the supervisory documents. We use this distribution as supervision to learn the 
corresponding conversation's topic distribution from the original text feature. In this way, the 
supervisory documents constrain the text features to focus on critical empathy information, thereby 
improving the accuracy of empathy prediction.

We conduct a series of experiments on the Multi-Modal Empathy Dataset in Counseling
\cite{zhu2023medic} (MEDIC) and the Mental Health Subreddits Dataset\cite{sharma2020computational}. 
To the best of our knowledge, the MEDIC stands as the only publicly 
available multi-modal empathy dataset. Specifically, we perform ablation experiments 
and then compare our method with related works on this multi-modal dataset. The 
experimental results show that our method improves the accuracy and F1 score with 
multi-modal input and assistance of supervised documents, compared to the single-modal and unassisted 
cases. Additionally, we conduct cross-dataset experiments on the Mental Health Subreddits 
Dataset\cite{sharma2020computational} to validate the generalization capabilities of our approach. Compared 
to related work, our method demonstrates clear advantages.

Contributions of this work are summarized as follows:

\textbf{(1)} We introduce a multi-modal empathy prediction network, which enables us to gather more comprehensive information across multiple modalities.

\textbf{(2)} We novelly leverage supervisory documents through a dedicated training framework to assist the text feature extraction. This integrates the rich empathy information from supervisory documents into the text features. 


\section{Related Work}
In this section we will give a brief review on previous works of empathy prediction and assisted 
training with privileged information.
\subsection{Empathy prediction} 
Empathy prediction is to estimate the level of empathy displayed in given communication scenarios, such as 
counseling sessions or online conversations. Related studies on predicting empathy mainly focus on developing 
network architectures to predict empathy. Gibson et al. \cite{gibson2016deep} introduced a recurrent 
neural network (RNN) that maps the text content of each speaker turn to a series of behavioral acts, 
which was then employed to initialize lower layers of the network to predict counselors' empathy 
ratings. They conducted the experiments on the dataset including motivational interviews collected from 
six independent clinical studies on addiction counseling. Sharma et al.
\cite{sharma2020computational} proposed a computational approach to comprehend the expression of 
empathy in online mental health platforms. They created a new unified, theoretically grounded 
framework to characterize empathy communication in text-based conversations. Xiao et al.
\cite{xiao2012analyzing} employed an N-gram language model based on a maximum 
likelihood strategy to differentiate empathic from non-empathic utterances to measure empathy. They 
carried out the experiments on the dataset of the transcripts from clinical studies on 
substance use. Kumano et al.\cite{kumano2011analyzing} developed a probabilistic model that employed a 
hierarchical structure involving facial expressions, and other behaviors, including utterances and 
gaze direction to estimate emotional interactions. Guda et al.\cite{guda2021empathbert} introduced 
EMPATH-BERT, a demographic-aware framework for empathy prediction based on Bidirectional 
Encoder Representations from Transformers\cite{devlin2018bert} (BERT). They used an
empathy-distress dataset including text pairs of news-responses for the empathy prediction.
Vasava et al.\cite{vasava-etal-2022-transformer} presented a framework that employs the RoBERTa model 
for training, followed by the addition of a few layers on top to fine-tune the transformer. They 
evaluated their results on the same dataset as that used by Guda et al.\cite{guda2021empathbert}.
Dey et al.\cite{dey2022enriching} performed a study on empathic language using a computational approach 
and linguistic analysis. They developed a novel system architecture to enhance the 
classification of empathy using state-of-the-art recurrent neural networks and transformer models. The 
experiments were conducted on a large corpus of 440 essays written by pre-med students, which narrated 
simulated patient-doctor interactions.

However, most predicting methods above only utilized unimodal input to predict 
empathy levels, especially text. This may lead to the loss of multi-modal information and 
the lack of diversity in feature representations. In this paper, we introduce an effective multi-modal 
empathy prediction network, which combines text, audio, and visual information, to predict empathy 
levels in multi-modal scenarios.

\subsection{Assisted training with privileged information} Assisted training with privileged information 
is to utilize additional information during the training process that may not be 
available during the testing phase. Privileged information can enhance the model's accuracy and 
efficiency by providing additional information. For instance, Vapnik et al.\cite{vapnik2009new} 
introduced the Learning Using Privileged Information (LUPI) paradigm. Unlike traditional machine learning 
methods, LUPI allowed the inclusion of privileged information during the training phase, which was not 
available during testing. They extended the Support Vector Machine (SVM) algorithm to create 
SVM+, which incorporated the privileged information. Wang et al. 
\cite{wang2007concept} proposed a novel ontology-assisted text document similarity measurement method 
called Concept Forest (CF). The CF approach constructed semantic representations of documents by mapping 
keywords to their respective synsets and forming a semantic tree using the WordNet lexical database. Li et 
al.\cite{li2011comment} introduced a novel Comment-Guided Learning (CGL) framework to bridge the knowledge 
gap between expert assessors and feature engineers in natural language processing (NLP) systems. The CGL 
framework utilizes comments from human assessors as privileged information during the training phase to 
guide feature encoding.
Hoeve et al.\cite{ter2020conversations} explored 
document-centered assistance in productivity contexts, addressing the gap in conversational digital 
assistants’ ability to facilitate document-related tasks. They developed reasonably accurate models to 
answer questions during conversations using document-centered assistance. 

We are inspired by the mentioned works and propose a supervisory documentation assisted
training framework that includes two components: a Latent Dirichlet Allocation (LDA) module and a 
topic distribution prediction module. The LDA module enables us to obtain the topic distribution of the 
supervision document, which we utilize as supervision to learn the topic distribution of the 
corresponding conversation from the text features. By constraining the text features through this 
assisted training framework, we can prioritize the content that is more relevant to empathy in the 
conversation, thereby improving the accuracy of empathy prediction.

\begin{figure*}[!ht]
\centering
\includegraphics[width=\linewidth]{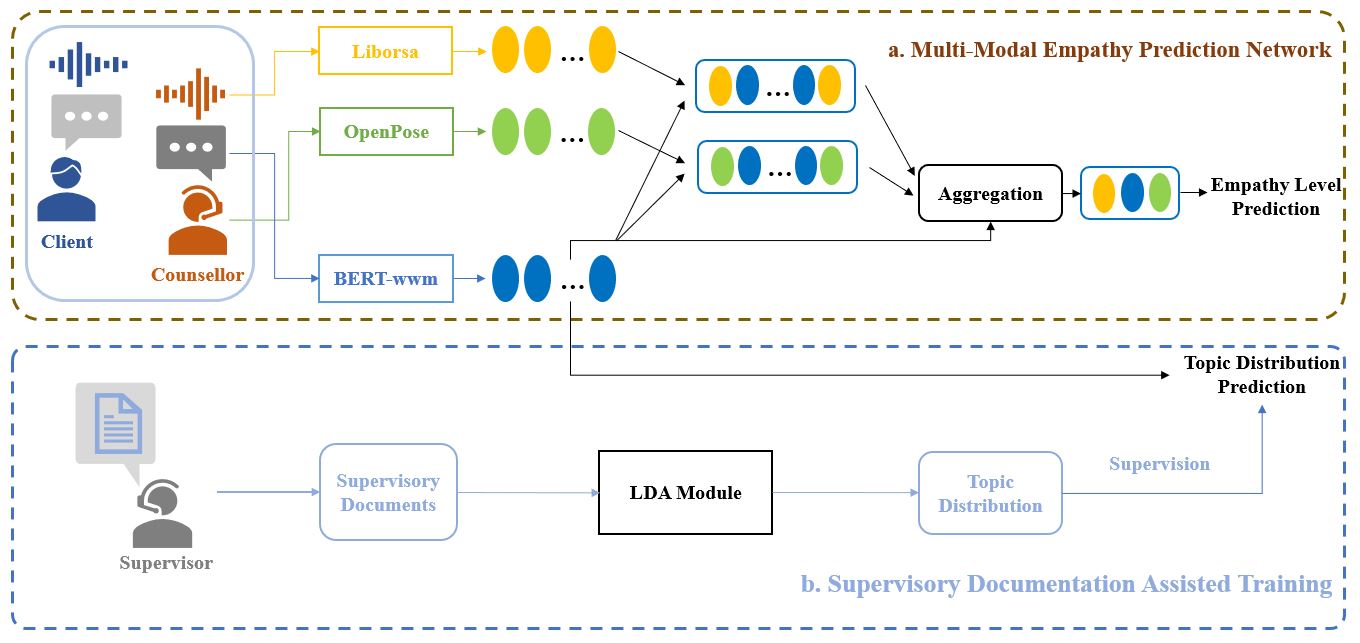}
\caption{The framework of Empathy Level Prediction in Multi-Modal Scenario with Supervisory Documentation Assistance method. It consists of two parts (a) Multi-Modal Empathy Prediction Network (b) Supervisory Documentation Assisted Training}
\label{fig:label_example}
\end{figure*}

\section{Problem Statement}
The original data comprises three types of information: text, audio, and visual information. 
It consists of $N$ conversation segments with corresponding text $\{T_i\}$, audio $\{A_i\}$, and video
 $\{V_i\}$ representations, as well as the associated supervisory documents $\{P_i\}$ in text form, 
where $i \in \{1, 2, \dots, N\}$. The training set is $S_{test}=\{T_{train}, A_{train}, V_{train}, 
P_{train}\}$ , where $T_{train}$, $A_{train}$, $V_{train}$, $P_{train}$ are the subsets of $\{T_i\}$, 
$\{A_i\}$, $\{V_i\}$ and $\{P_i\}$ respectively. 
The testing set is $S_{test}=\{T_{test}, A_{test}, V_{test}\}$. 
In addition, the supervisory documents are only available in the training set, 
indicating that no additional information can be utilized during testing. Our goal is to utilize 
the rich semantic information in the multi-modal representations and supervisory documents to 
predict the levels of empathy.

\section{Proposed Method}
As illustrated in Figure 2, we introduce a multi-modal empathy prediction network assisted by 
supervisory documentation to predict empathy under the multi-modal scenario. Our network comprises two 
primary components: a multi-modal empathy prediction network, which combines text, audio, and visual 
features to predict empathy labels, and the supervisory documentation assisted training process that 
leverages supervisory documents to learn the distribution of topics, which is then used to supervise the 
prediction of dialogue's topic distribution.

The multi-modal empathy prediction network consists of three steps: feature extraction, cross-modal 
combination and multi-modal aggregation. In the feature extraction part, we use three pre-trained
models to extract the feature representations of the source data of three modalities. In the cross-modal 
combination part, we employ a combination similar to the attention mechanism on the extracted unimodal 
features to learn the bimodal fusion between text modality and other modalities. In the multi-modal aggregation part, we further concatenate the bimodal
fusions and text feature using a Long Short-Term Memory (LSTM) layer to deeply fuse the three modalities, because previous 
studies\cite{mai2019divide} have demonstrated that text modality often plays a major role among the 
three modalities. Thus we obtain the final deeply-fused feature representations of the three modalities, which are input to a softmax layer for predicting the final empathy prediction.

Then we present a supervisory documentation-assisted training process, which comprises 
two components: a Latent Dirichlet Allocation (LDA) module and a topic distribution prediction module. 
The LDA module learns the topic distribution from the supervisory documents and the probabilities of 
words associated with each topic, which enables us to obtain the topic distribution for a supervisory 
document in the training set. The topic distribution prediction module utilizes a linear layer with a 
softmax function applied to the text feature to predict the topic distribution of the conversation, 
using the topic distribution of the corresponding supervisory document as the supervision, instead of 
directly using the supervisory documents.

\subsection{Multi-Modal Empathy Prediction Network}
The inconsistency in multi-modal representations presents a significant challenge. To tackle these
problems, we develop a multi-modal empathy prediction network that comprises three modules: a 
feature extraction module, a cross-modal combination module, and a multi-modal aggregation module. The 
network predicts empathy labels by the multi-modal fusion, which is aggregated by the 
bimodal features and the text feature. This allows for a more accurate prediction of empathy labels.

\subsubsection{Feature Extraction} Initially, we obtain the original multi-modal data from the training 
set, labeled as $T_{train}$, $A_{train}$, and $V_{train}$, corresponding to the text, audio, and vision 
modality, respectively. Subsequently, we model the original data from each modality into features 
utilizing three pre-trained networks - Librosa\cite{mcfee2015librosa}, OpenPose\cite{cao2019openpose} 
and BERT-wwm\cite{cui2021pre}. These methods have been widely used for extracting Mel-frequency Cepstral 
Coefficients (MFCC) features from audio, body posture keypoint features, and text features. The hidden 
state outputs of the three modalities are represented as $K_T$, $K_A$, and 
$K_V$, respectively, as indicated in Equation (1). These preliminary features will serve as 
the inputs for the subsequent cross-modal combination module for more profound fusion.
\begin{equation}
\begin{aligned}
    K_A = Librosa[A_{train}] \\
    K_V = OpenPose[V_{train}] \\
    K_T = BERT_{wwm}[T_{train}]
\end{aligned}
\end{equation}

\subsubsection{Cross-modal Combination} In the cross-modal fusion section, we employ an approach similar to 
the attention mechanism\cite{vaswani2017attention} to design bimodal feature representations between the 
text modality and other modalities. The module takes the unimodal features $K_T$, $K_A$, 
and $K_V$ as inputs, and produces bimodal representations $C_{T\& V}$ and $C_{T\& A}$ as shallow fusions 
of the multi-modal information. Initially, we employ a non-linear projection layer to align the text 
feature's dimension with those of the other modalities, as presented in Equation (2).
\begin{equation}
\setlength{\abovedisplayskip}{15pt}
\setlength{\belowdisplayskip}{15pt}
    K_T^{'} = tanh(K_T \mathbf{W_T} + b_T) 
\end{equation}

Next, we use a covariance matrix $K_V (K_T{'})^{T}$ to compute affinity scores for all hidden state 
pairs of the corresponding modalities. To generate attention weights $A_V$ for each timestep of the 
visual (or audio) features across the text, we apply the softmax function to normalize the covariance 
matrix row-wise. Similarly, attention weights $A_T$ for each word across the visual (or audio) features 
are obtained by normalizing the covariance matrix column-wise. The attention contexts of text features 
can be computed by taking the product of the attention weights $A_V$ and the new text feature $K_T^{'}$. 
Finally, we get the bimodal feature $C_{T\& V}$ by multiplying the concatenated matrix of the visual 
feature $K_V$ and the attention contexts $C_V$ by the attention weight $A_T$. Similarly, we obtain 
another bimodal feature $C_{T\& A}$ by conducting the same process on the text and audio features.
These bimodal features serve as the inputs for the following multi-modal aggregation module to achieve 
the final fusion. The whole process of generating bimodal features is shown in the equations below:

\begin{equation}
\begin{aligned}
    & A_V = softmax(K_V (K_T^{'})^{T}) \\
    & A_T = softmax(K_T^{'} K_V^{T}) \\
    & C_V = A_V K_T^{'} \\
    & C_{T\& V} = A_T \times Concat[K_V, C_V]
\end{aligned}
\end{equation}

\subsubsection{Multi-Modal Aggregation} We define $C_{T \&V}$ and $C_{T \&A}$ as co-dependent representations 
of the text and vision (or audio) modalities. However, these bimodal fusions have limitations: they are 
shallow fusions and do not effectively capture the knowledge in the text modality, nor do they support 
trimodal fusion. To overcome this limitation, we propose a multi-modal aggregation module that 
aggregates the bimodal fusions and the text features through an LSTM layer to obtain the final 
multi-modal representation, $L_{agg}$. This representation plays an important role in predicting 
empathy labels. The equation for this module is presented below:
\begin{equation}
    L_{agg} = LSTM([C_{T\& V}, C_{T\& A}, K_T])
\end{equation}

The multi-modal empathy prediction network takes the original multi-modal data $T_{train}$, $A_{train}$, 
and $V_{train}$ as inputs and produces the deeply-fused multi-modal representation $L_{agg}$ for empathy 
prediction. In this network, we take full advantage of the multi-modal information and prioritize the 
text modality, which plays a more significant role than other modalities during the fusion process. 
Consequently, compared to related works on empathy prediction with unimodal information, our proposed 
network can learn empathetic knowledge from multi-modal features, resulting in improved accuracy in 
assessing empathy levels.

\subsection{Supervisory Documentation Assisted Training}
The additional challenge faced by this issue lies in optimizing the fusion of multi-modal 
features while incorporating supervisory documents. The content of supervisory documents is 
often verbose and may contain irrelevant information. Directly converting such text into 
feature vectors not only significantly increases computational complexity but also risks 
contaminating the original feature vectors, leading to a substantial decrease in the accuracy of 
empathy prediction. To address the limitations, we propose a supervisory documentation assisted 
training network comprising two parts: the Latent Dirichlet Allocation (LDA) module and a topic 
distribution prediction module. This network enables the learning of the topic distribution of 
supervisory documents, which is then utilized to supervise the learning of the dialogue's topic 
distribution.

\subsubsection{Latent Dirichlet Allocation} We apply the Latent Dirichlet Allocation (LDA) module on the supervisory documents to identify potential topics with their distributions. Latent Dirichlet Allocation 
is a statistical model for uncovering latent topics that are present in a collection of documents. 
It is based on the assumption that a document is a mixture of several topics, and each topic is 
represented by a distribution of words. The goal of LDA is to learn the underlying topic distribution of 
a collection of documents, as well as the word distribution of each topic. The model is unsupervised, 
meaning it can automatically discover latent topics without prior knowledge or human intervention.

Specifically, we begin by setting the number of topics, denoted as $K$, that each document in the 
training set $P_{train}$ will be represented by a K-dimensional probability vector. Subsequently, the 
LDA module randomly assigns each word in each document to one of the topics and calculates the 
likelihood of the data given the current topic assignments. And then, the algorithm updates the 
topic distribution for each document and the word distribution for each topic. This iterative process 
continues until the algorithm converges to a stable set of topic distributions and word distributions.

As a result of the LDA module, we obtain two sets of parameters: the word probabilities 
$\{ w_1, w_2 , \dots, w_K \}$ given a topic 
and the topic probabilities $( p_{i1}, p_{i2}, \dots, p_{iK} )$ given a document $T_i$. By utilizing 
the latter set of parameters, we can 
represent each document and map any new document into the topic space through the LDA module. This 
allows us to express each document as a probability distribution of the topic, which yields a 
K-dimensional probability vector of topic distribution. These results can be used for topic 
classification tasks in the subsequent module and represented in Equation (5).
\begin{equation}
\begin{aligned}
    & Topics = \{ w_1, w_2 , \dots, w_K \} \\
    & y_{dis}(T_i) = ( p_{i1}, p_{i2}, \dots, p_{iK} ) \\
    & Where \quad \sum\limits_{j=1}^{K} p_{ij} = 1, \quad 0 \leq p_{ij} \leq 1
\end{aligned}
\end{equation}

\subsubsection{Topic Distribution Prediction} After applying the LDA module to extract the topic distribution, 
we propose a linear model that incorporates softmax layers on the text feature $K_T$ to predict 
the distribution of topics within a conversation. Specifically, each conversation $T_i$ is associated 
with a corresponding supervisory document $P_i$. The conversation $T_i$ possesses a topic distribution of 
$(p_{i1}, p_{i2}, \dots, p_{iK})$. In this way, the topic distribution serves as a supervision mechanism to 
guide the learning process on the text features $K_T$. The module is formally presented through Equation 
(6).
\begin{equation}
\begin{aligned}
    & z_{dis} =  K_T^{'} \mathbf{W_{dis}} + b_{dis} \\
    & \hat{y}_{dis} = softmax (z_{dis}) = \frac{exp(z_{dis})}{\sum\limits_{j=1}^{K} exp(z_j)}
\end{aligned}
\end{equation}
where $\hat{y}_{dis}$ refers to the final prediction of the distribution of the conversations.

The supervisory documentation assisted training network learns the distribution of supervisory documents 
through the Latent Dirichlet Allocation module. Then the network uses text features to predict the 
conversation distribution with the supervisory document distribution serving as supervision. This 
approach enhances the understanding of empathetic expressions by supervising the text features and 
increasing their attention to the essential content for empathy patterns of the text. 

\subsection{Loss Function}
The loss function of the proposed multi-modal fusion model with a supervisory documentation-assisted 
network consists of two parts. The first part is the loss function of the multi-modal empathy prediction 
network. In the previous subsection, we explained how the model generates the final multi-modal 
representation, $L_{agg}$. The model then feeds this representation to a linear layer with a softmax 
function to obtain the empathy label predictions, as shown in the following equations:
\begin{equation}
\begin{aligned}
    & z_{emp} = L_{agg} \mathbf{W_{emp}} + b_{emp} \\
    & \hat{y}_{emp} = softmax (z_{emp})
\end{aligned}
\end{equation} 
$\hat{y}_{emp}$ refers to the predicted empathy labels. Then we utilize the cross-entropy loss 
function $L_s$ on the given true label $y_{i}$, as presented in the equation below:
\begin{equation}
    L_s = - \frac{1}{N_{train}} \sum\limits_{i} \hat{y}_{emp,i}log(y_{i})
\end{equation} where $N_{train}$ denotes the size of the training set.

In the second part, we employ the KL divergence loss $L_t$ to evaluate the distance of 
the predicted distribution $\hat{y}_{dis}$ and the true distribution $y_{dis}$, which is demonstrated
in the following equation.
\begin{equation}
    L_t = \sum\limits_{i} \hat{y}_{dis}(T_i) log \frac{\hat{y}_{dis}(T_i)}{y_{dis}(T_i)}
\end{equation}

Finally, we obtain the total loss function $L$ by linearly combining $L_s$ and $L_t$ with the 
corresponding weights $w_s$ and $w_t$, as shown in the equation below:
\begin{equation}
    L = w_s L_s + w_t L_t
\end{equation}

To provide a clearer illustration of the method's workflow, we present a detailed training process  
in Algorithm 1.

\begin{table*}[t]
\centering
\caption{The performance of our model using unimodal, 
bimodal and multi-modal features}
\resizebox{\linewidth}{!}{
  \begin{tabular}{>{\centering\arraybackslash}p{\dimexpr(\linewidth-2\tabcolsep)/4\relax}|>{\centering\arraybackslash}p{\dimexpr(\linewidth-2\tabcolsep)/4\relax}|>{\centering\arraybackslash}p{\dimexpr(\linewidth-2\tabcolsep)/12\relax}>
  {\centering\arraybackslash}p{\dimexpr(\linewidth-2\tabcolsep)/12\relax}|>
  {\centering\arraybackslash}p{\dimexpr(\linewidth-2\tabcolsep)/12\relax}>
  {\centering\arraybackslash}p{\dimexpr(\linewidth-2\tabcolsep)/12\relax}|>
  {\centering\arraybackslash}p{\dimexpr(\linewidth-2\tabcolsep)/12\relax}>
  {\centering\arraybackslash}p{\dimexpr(\linewidth-2\tabcolsep)/12\relax}}
    \toprule[1.2pt]
    \multirow{2}{*}{\textbf{Model}} & \multirow{2}{*}{\textbf{Source}} & \multicolumn{2}{c|}{\textbf{EE label}} & \multicolumn{2}{c|}{\textbf{ER label}} & \multicolumn{2}{c}{\textbf{CR label}} \\
    & & \textbf{Acc.}  & \textbf{F1}   & \textbf{Acc.} & \textbf{F1}  & \textbf{Acc.}   & \textbf{F1} \\
    \midrule
    \multirow{3}{*}{\textbf{Unimodal}} & Text & $0.753$ & $0.732$ & $0.783$ & $0.773$ & $0.750$ & $0.754$\\
    & Audio & $0.721$ & $0.719$ & $0.783$ & $0.781$ & $0.651$ & $0.653$\\
    & Video & $0.656$ & $0.659$ & $0.691$ & $0.685$ & $0.586$ & $0.519$\\
    \midrule
     \multirow{3}{*}{\textbf{Bimodal}} & Text+Audio & $0.773$ & $0.776$ & $0.783$ & $0.775$ & $0.783$ & $0.778$\\
     & Text+Video & $0.857$ & $0.858$ & $0.796$ & $0.792$ & $0.822$ & $0.823$ \\
     & Audio+Video & $0.753$ & $0.758$ & $0.789$ & $0.783$ & $0.763$ & $0.758$ \\
     \midrule
     \multirow{2}{*}{\textbf{Multi-Modal}} & \textbf{Text+Audio} & \multirow{2}{*}{$\textbf{0.883}$} & \multirow{2}{*}{$\textbf{0.884}$} & \multirow{2}{*}{$\textbf{0.822}$} & \multirow{2}{*}{$\textbf{0.812}$} & \multirow{2}{*}{$\textbf{0.829}$} & \multirow{2}{*}{$\textbf{0.827}$} \\
     & \mbox{\textbf{+Video}} & & & & & & \\
  \bottomrule[1.2pt]
  \end{tabular}
  }
  \label{tab:freq}
\end{table*}

\begin{algorithm}[ht]
	\caption{The learning algorithm of the proposed method.}
	\label{algorithm1}
	\begin{algorithmic}[1]
		\Statex $\begin{aligned}[t]
            \textbf{Input:} &\quad \text{Training set:} \, \mathcal{S}_{Train}=\{T_i, A_i, V_i, P_i, y_i\}_{i=1}^{N}; \\
                            &\quad \text{Number of topics:} \, K; \\
                            &\quad \text{Number of iterations:} \, T
            \end{aligned}$
		\Statex \textbf{Output:} Prediction of conversation's empathy level $\hat{y}$
        \State Obtain the feature representations of three modalities, namely $K_A$, $K_V$, and $K_T$. 
        \State Get the topic distribution $\theta$ of the supervisory documents.
        $y_{dis} = LDA [\{P_i\}_{i=1}^{N}, K ]$
        \State Initialize the weight matrices $\mathbf{W_{T}}$, $\mathbf{W_{dis}}$, $\mathbf{W_{egg}}$ and the bias matrices $b_{T}$, $b_{dis}$, $b_{emp}$ randomly
		\For {iterations $i=1$ to $T$}
        \State Calculate the aligned text features $K_T^{'}$.
        \State Get the multi-modal representations $L_{agg}$.
		\State Calculate the predicted empathy labels $\hat{y}_{emp}$.
        \State Obtain the loss of the predicted empathy labels $L_s$.
        \State Calculate the predicted distributions $\hat{y}_{dis}$.
		\State Obtain the KL divergence loss of the predicted distributions $L_t$.
        \State Update the weight matrices $\mathbf{W_{T}}$, $\mathbf{W_{dis}}$, $\mathbf{W_{egg}}$ and the bias matrices $b_{T}$, $b_{dis}$, $b_{emp}$ with the loss.
		\EndFor
        \State Calculate the multi-modal representations $L_{agg}$
        \State Calculate the final predicted empathy labels 
        \Statex \hspace{1em} $\hat{y} = argmax( softmax (L_{agg} \mathbf{W_{emp}} + b_{emp})$ \\
        \Return{The prediction $\hat{y}$}
	\end{algorithmic}
\end{algorithm}

\section{Experiments}

\subsection{Experimental Setup}
\subsubsection{Datasets} To evaluate our proposed method, we conduct experiments on the Multi-Modal 
Empathy Dataset in Counseling\cite{zhu2023medic} (MEDIC), which is currently the only known public 
multi-modal empathy dataset in counseling scenarios. 
This dataset is built on wild, face-to-face, authentic counseling sessions. It comprises 771 
video clips including the text modality reflecting the conversations' content, the visual modality that 
captures the body and facial expressions, and the audio modality recording the voices. To evaluate the 
degree of empathy between the counselors and their clients, the dataset proposes three labels: 
expression of experience (EE), emotional reaction (ER), and cognitive reaction (CR). The empathy labels were 
annotated by trained annotators.  We obtained the supervisory documents from the 
source of the dataset, UM Psychology’s counseling case courses. We discovered that for each original mental 
counseling session video in the dataset, there is an accompanying text segment providing supervision and 
commentary on the empathy process by external supervisors. These contents can be used as supplementary data 
to the primary dataset, as they contain rich empathy-related information that can be leveraged to predict 
empathy labels. Following the work of the dataset paper\cite{zhu2023medic}, we divided MEDIC into three 
parts, consisting of a training set, a validation set, and a testing set, with a ratio of 7:1:2.

To further investigate the generalization capabilities of our proposed model, we conducted cross-dataset 
experiments 
with the trained network on the Mental Health Subreddits Dataset\cite{sharma2020computational}. The dataset
comprises 3,081 pairs of psychological question-and-answer dialogues collected from Reddit, annotated with 
three empathy indicators similar to those in the MEDIC dataset, namely Emotional Reactions, Explorations, 
and Interpretations. Due to the lack of multi-modal empathy datasets, to the best of our knowledge, this is 
the only text-based empathy dataset for psychological counseling dialogues with annotations comparable to 
the MEDIC dataset. Consequently, we were able to directly transfer our trained ablation model, which solely 
relies on textual input, onto the Mental Health Subreddits dataset without the necessity for retraining. 
Following their work\cite{sharma2020computational}, we split the dataset into training, validation, and test sets with a ratio of 75:5:20 
to align with comparable methods.

\noindent
\subsubsection{Implementation Details} The proposed model was trained using a dual setup of 3090 
graphics cards. A batch size of 32 was used for training, along with a learning rate of 1e-3 and a 
dropout rate of 0.3. The proposed model underwent 60 epochs of training. The number of topics 
$K$ is set to 10 using the grid search. The weighting factors for the loss functions, denoted as $w_s$ and 
$w_t$, are assigned values of 0.84 and 0.16, respectively. 
We computed accuracy and weighted F1 scores based on the true and predicted labels to assess the model's 
performance. 
Higher accuracy and F1 score values indicate better model performance.

\subsection{Ablation Study and Analysis}
As outlined in Section 4, our framework consists of two major components. To assess the 
effectiveness of each component, we conducted ablation studies. We utilized similar networks 
using unimodal, bimodal, and multi-modal features to assess the performance of the multi-modal 
fusion network. Additionally, we used a similar network without the supervisory documentation-
assisted training network to determine the efficacy of supervisory documents. Detailed results 
are presented in Tables 1 and 2.

\begin{table*}[t]
\centering
\caption{Comparison between the models with and without supervisory documentation assisted training}
\resizebox{\linewidth}{!}{
  \begin{tabular}{>{\centering\arraybackslash}p{\dimexpr(\linewidth-2\tabcolsep)/7\relax}|>{\centering\arraybackslash}p{\dimexpr(\linewidth-2\tabcolsep)/7\relax}>{\centering\arraybackslash}p{\dimexpr(\linewidth-2\tabcolsep)/7\relax}|>
  {\centering\arraybackslash}p{\dimexpr(\linewidth-2\tabcolsep)/7\relax}>
  {\centering\arraybackslash}p{\dimexpr(\linewidth-2\tabcolsep)/7\relax}|>
  {\centering\arraybackslash}p{\dimexpr(\linewidth-2\tabcolsep)/7\relax}>
  {\centering\arraybackslash}p{\dimexpr(\linewidth-2\tabcolsep)/7\relax}}
    \toprule[1.2pt]
    \multirow{2}{*}{\textbf{Model}} & \multicolumn{2}{c|}{\textbf{EE label}} & \multicolumn{2}{c|}{\textbf{ER label}} & \multicolumn{2}{c}{\textbf{CR label}} \\
    & \textbf{Acc.}  & \textbf{F1}   & \textbf{Acc.} & \textbf{F1}  & \textbf{Acc.}   & \textbf{F1} \\
    \midrule
    Model & \multirow{3}{*}{$0.851$} & \multirow{3}{*}{$0.852$} & \multirow{3}{*}{$0.783$}  & \multirow{3}{*}{$0.779$}  & \multirow{3}{*}{$0.796$}   & \multirow{3}{*}{$0.788$} \\
    without & & & & & & \\
    SDAT & & & & & & \\
    \textbf{Model} & \multirow{3}{*}{$\textbf{0.883}$} & \multirow{3}{*}{$\textbf{0.884}$} & \multirow{3}{*}{$\textbf{0.822}$} & \multirow{3}{*}{$\textbf{0.812}$} & \multirow{3}{*}{$\textbf{0.829}$} & \multirow{3}{*}{$\textbf{0.827}$} \\
    \textbf{with} & & & & & &\\
    \textbf{SDAT} & & & & & &\\
  \bottomrule[1.2pt]
  \end{tabular}
  }
  \label{tab:freq}
\end{table*}

\begin{table*}[t]
\centering
\caption{Comparison with related works}
\resizebox{\linewidth}{!}{
  \begin{tabular}{>{\centering\arraybackslash}p{\dimexpr(\linewidth-2\tabcolsep)/7\relax}|>{\centering\arraybackslash}p{\dimexpr(\linewidth-2\tabcolsep)/7\relax}>{\centering\arraybackslash}p{\dimexpr(\linewidth-2\tabcolsep)/7\relax}|>
  {\centering\arraybackslash}p{\dimexpr(\linewidth-2\tabcolsep)/7\relax}>
  {\centering\arraybackslash}p{\dimexpr(\linewidth-2\tabcolsep)/7\relax}|>
  {\centering\arraybackslash}p{\dimexpr(\linewidth-2\tabcolsep)/7\relax}>
  {\centering\arraybackslash}p{\dimexpr(\linewidth-2\tabcolsep)/7\relax}}
    \toprule[1.2pt]
    \multirow{2}{*}{\textbf{Model}} & \multicolumn{2}{c|}{\textbf{EE label}} & \multicolumn{2}{c|}{\textbf{ER label}} & \multicolumn{2}{c}{\textbf{CR label}} \\
    & \textbf{Acc.}  & \textbf{F1}   & \textbf{Acc.} & \textbf{F1}  & \textbf{Acc.}   & \textbf{F1} \\
    \midrule
    Concat & $0.819$ & $0.810$ & $0.703$ & $0.699$ & $0.735$ & $0.731$\\
    TFN & $0.758$  & $0.758$ & $0.729$  & $0.719$ & $0.722$  & $0.712$\\
    SWAFN & $0.864$ & $0.863$ & $0.743$ & $0.743$ & $0.783$ & $0.785$\\
    BERT & \multirow{2}{*}{$0.727$} & \multirow{2}{*}{$0.722$} & \multirow{2}{*}{$0.757$} & 
    \multirow{2}{*}{$0.730$} & \multirow{2}{*}{$0.776$} & \multirow{2}{*}{$0.762$}\\
    {-based} & & & & & &\\
    \midrule
    \textbf{Ours} & $\textbf{0.883}$ & $\textbf{0.884}$ & $\textbf{0.822}$ & $\textbf{0.812}$ & $\textbf{0.829}$ & $\textbf{0.827}$\\
  \bottomrule[1.2pt]
  \end{tabular}
    }
  \label{tab:freq}
\end{table*}

\subsubsection{Ablation Study on Multi-Modality} In our multi-modality ablation experiments, we utilized 
similar networks that employ unimodal, bimodal, and multi-modal features to assess the performance 
of the multi-modal empathy prediction network. The results of the experiments are presented in Table 1, 
indicating that the model achieves the best performance with multi-modal input on each label. This 
outcome suggests that the model learns significantly from the multi-modal information. Compared to 
the model's accuracy with unimodal input, our model demonstrates conspicuous improvements of 
17.3\%, 7.0\%, and 16.6\% on each label. The model's most significant enhancement is observed in 
the EE label, indicating that the multi-modal empathy prediction network effectively captures and 
utilizes empathy information from audio and visual features, derived from the client's sounds and 
movements, to predict the empathy level. Conversely, the improvements on ER and CR labels are less 
pronounced because mental counselors usually regulate their emotional expression during counseling 
sessions. Additionally, our findings indicate that the bimodal input accuracy and F1 score are 
higher than those of the model with unimodal input. This outcome demonstrates that bimodal input 
has more empathy information for prediction than unimodal input.

\subsubsection{Ablation Study on Supervisory Documents} Table 2 presents the results of our ablation 
study on supervisory documents, where we compared models with and without the supervisory 
documentation assisted training (SDAT) module. Our findings indicate that the inclusion of supervisory 
documents improves accuracy for each label by 3.2\%, 3.9\%, and 3.3\%, respectively. These results 
demonstrate that our model learns valuable empathy information from the supervisory documents. 
Notably, we observed a slightly higher improvement in the ER label compared to other labels. 
This result can be attributed to the fact that supervisory documents mainly focus on evaluating 
the emotional reactions of mental counselors during counseling sessions, which provides crucial 
information for predicting the ER label.

\subsection{Comparisons to Related Works}
In the section of comparisons to the related works, we compared our approach with baselines on the dataset 
paper\cite{zhu2023medic} and a BERT-based empathy prediction model. 
The straightforward multi-modal concatenation model, TFN model, and SWAFN model serve as the baselines for the 
MEDIC dataset. Detailed descriptions of these methods can be found in the dataset's original paper. 
In addition, we conducted comparative 
experiments using the empathy prediction model proposed in the paper\cite{sharma2020computational} against 
our model. This model is a variant of BERT\cite{devlin2018bert}, which implies that it exclusively utilizes 
the textual information from the dataset. To the best of our knowledge, it is the only recent empathy prediction work with publicly available source code.

The experimental findings are presented in Table 3. Our proposed approach surpasses 
prior related works, exhibiting the highest accuracy levels of 88.3\%, 82.2\%, and 82.9\% for 
empathy labels, indicating improvements of 1.9\%, 6.5\%, and 4.6\% compared to the best-
performing method for each label. Additionally, our model achieved the highest F1 score of 0.884, 
0.812, and 0.827. Notably, the improvements are more significant for the ER and CR labels, 
suggesting our method's superiority over other approaches for mental counselor-related labels. 
This advantage can be attributed to the evaluation of the supervisor to the treatment process of the 
mental counselor, which is a central aspect of the supervisory documents and immensely beneficial 
in predicting ER and CR labels.

\subsection{Cross-Dataset Experiments}

\begin{table*}[t]
\centering
\caption{Cross-Dataset Performance}
\resizebox{\linewidth}{!}{
  \begin{tabular}{>{\centering\arraybackslash}p{\dimexpr(\linewidth-2\tabcolsep)/7\relax}|>{\centering\arraybackslash}p{\dimexpr(\linewidth-2\tabcolsep)/7\relax}>{\centering\arraybackslash}p{\dimexpr(\linewidth-2\tabcolsep)/7\relax}|>
  {\centering\arraybackslash}p{\dimexpr(\linewidth-2\tabcolsep)/7\relax}>
  {\centering\arraybackslash}p{\dimexpr(\linewidth-2\tabcolsep)/7\relax}|>
  {\centering\arraybackslash}p{\dimexpr(\linewidth-2\tabcolsep)/7\relax}>
  {\centering\arraybackslash}p{\dimexpr(\linewidth-2\tabcolsep)/7\relax}}
    \toprule[1.2pt]
    \multirow{2}{*}{\textbf{Model}} & \multicolumn{2}{c|}{\textbf{Emotion Reactions}} & \multicolumn{2}{c|}{\textbf{Explorations}} & \multicolumn{2}{c}{\textbf{Interpretations}} \\
    & \textbf{Acc.}  & \textbf{F1}   & \textbf{Acc.} & \textbf{F1}  & \textbf{Acc.}   & \textbf{F1} \\
    \midrule
    Log. Reg. & $41.69$ & $42.69$ & $67.08$ & $46.63$ & $70.58$ & $49.77$ \\
    RNN & $71.63$ & $42.85$ & $85.58$ & $30.74$ & $76.21$ & $51.76$ \\
    HRED & $71.11$ & $44.10$ & $85.58$ & $30.74$ & $79.65$ & $54.16$ \\
    BERT & $72.13$ & $50.41$ & $89.35$ & $56.54$ & $82.16$ & $61.20$ \\
    GPT-2 & $76.69$ & $71.65$ & $88.25$ & $58.28$ & $82.32$ & $62.27$ \\
    DialoGPT & $66.07$ & $51.66$ & $89.65$ & $70.65$ & $81.85$ & $68.95$ \\
    RoBERTa & $76.99$ & $70.35$ & $90.58$ & $63.41$ & $82.16$ & $61.38$ \\
    Baseline  & $\textbf{79.43}$ & $\textbf{74.46}$ & $\textbf{92.61}$ & $72.58$ & $\textbf{84.04}$ & $62.60$ \\
    \midrule
    \textbf{Ours} & $73.70$ & $69.66$ & $78.50$ & $\textbf{78.52}$ & $72.21$ & $\textbf{70.83}$\\
  \bottomrule[1.2pt]
  \end{tabular}
    }
  \label{tab:freq}
\end{table*}

\begin{table}[t]
\centering
\caption{Misclassified Examples of EE label}
\resizebox{\linewidth}{!}{
  \begin{tabular}{>{\centering\arraybackslash}p{\dimexpr(\linewidth*5/7-2\tabcolsep)/1\relax}|>{\centering\arraybackslash}p{\dimexpr(\linewidth-2\tabcolsep)/7\relax}>{\centering\arraybackslash}p{\dimexpr(\linewidth-2\tabcolsep)/7\relax}}
    \toprule[1.2pt]
    \textbf{Conversations} & \textbf{True} & \textbf{Predicted}\\
    \midrule
    \textbf{Client:} What else can I do!& \multirow{4}{*}{$1$} & \multirow{4}{*}{$0$} \\
    \textbf{Mental Counselor:} It’s because when faced with & & \\
     a situation that seems to have a choice, you seem & & \\
     to have no choice. & & \\
    \midrule
    \textbf{Client:} Worry about their relationship? & \multirow{4}{*}{$1$} & \multirow{4}{*}{$0$}\\
    \textbf{Mental Counselor:} It seems that you & & \\
    have an intolerance for inequality that& & \\
    disturbs you. & & \\
  \bottomrule[1.2pt]
  \end{tabular}
  }
  \label{tab:freq}
\end{table}

\begin{table}[t]
\centering
\caption{Misclassified Examples of ER label}
\resizebox{\linewidth}{!}{
  \begin{tabular}{>{\centering\arraybackslash}p{\dimexpr(\linewidth*5/7-2\tabcolsep)/1\relax}|>{\centering\arraybackslash}p{\dimexpr(\linewidth-2\tabcolsep)/7\relax}>{\centering\arraybackslash}p{\dimexpr(\linewidth-2\tabcolsep)/7\relax}}
    \toprule[1.2pt]
    \textbf{Conversations} & \textbf{True} & \textbf{Predicted}\\
    \midrule
    \textbf{Client:} Yes, that is other people's business, but why, when it comes to me, I don't know what's wrong with me, that is, why can't I be the kind of person who has the passion and motivation to build a big family and fill the home with laughter every day. & \multirow{12}{*}{$1$} & \multirow{12}{*}{$0$} \\
    \textbf{Mental Counselor:} It sounds like you really want such a family, but it may be a conflict. You really want it, but I think that kind of desire is too far away from you, so on the other hand, you may feel that it is someone else's family and has nothing to do with you at all. & & \\
    \midrule
    \textbf{Client:} Fine. & \multirow{7}{*}{$1$} & \multirow{7}{*}{$0$}\\
    \textbf{Mental Counselor:} Well, if that's the case, then & & \\
    maybe, um, there will be a choice here. Do you & & \\
    choose a person who is like a big tree and can & & \\
    give you enough care when you need it, or do you & & \\
    need someone who is similar to you but can & & \\ 
    accompany you to move forward? & & \\
  \bottomrule[1.2pt]
  \end{tabular}
  }
  \label{tab:freq}
\end{table}

We conducted comparative analyses with existing methods on the  Mental Health Subreddits Dataset
\cite{sharma2020computational} dataset, with experimental results presented in Table 4. Since this dataset 
contains only textual dialogue information, we directly applied the pre-trained text ablation model to generate 
the corresponding empathy labels without requiring additional training.

We find that our model achieves a performance that ranks in the upper middle range 
across all models, with an accuracy that lags by an average of 10.6\% behind the optimal model. 
However, in terms of the F1 score, our model surpasses the optimal model on the Explorations and 
Interpretations labels. Given that our model was not trained on this 
dataset, the discrepancy in accuracy falls within a desirable range. Moreover, our model exhibits a small 
discrepancy between accuracy and the F1 score, with the F1 score outperforming the majority of models.
This suggests that the model exhibits a well-balanced performance, capable of effectively identifying 
categories with fewer instances of labels, thereby demonstrating commendable overall efficacy.

\subsection{Error Analysis}
As presented in Table 5, we identified two random cases of misclassified EE labels that occurred when 
the client's question was relatively straightforward. In these cases, the model did not infer 
any expression of experience in the conversation. However, certain punctuation marks at the end of the 
sentences altered the emotional tone conveyed by the client. For example, in the first dialogue in Table 5, the phrase ``What else can I do" is neutral in tone, but with the addition of an exclamation mark, we can 
sense the client's urgency. These punctuation marks were overlooked during the preprocessing phase of our 
work, potentially leading to misclassification. These examples highlight the need to pay closer attention 
to emotional punctuation in future research to improve our results.

\begin{table}[t]
\centering
\caption{Misclassified Examples of CR label}
\resizebox{\linewidth}{!}{
  \begin{tabular}{>{\centering\arraybackslash}p{\dimexpr(\linewidth*5/7-2\tabcolsep)/1\relax}|>{\centering\arraybackslash}p{\dimexpr(\linewidth-2\tabcolsep)/7\relax}>{\centering\arraybackslash}p{\dimexpr(\linewidth-2\tabcolsep)/7\relax}}
    \toprule[1.2pt]
    \textbf{Conversations} & \textbf{True} & \textbf{Predicted}\\
    \midrule
    \textbf{Client:} Hmm. Could you repeat that?& \multirow{7}{*}{$2$} & \multirow{7}{*}{$1$} \\
    I didn't quite catch you. & & \\
    \textbf{Mental Counselor:} Yeah, so, like, 
    when you were talking earlier about 
    changing jobs, it got me thinking about 
    how every different job switch is 
    unpredictable, especially when you're 
    quitting and starting a new one. You 
    know what I mean? & & \\
    \midrule
    \textbf{Client:} Did I do something wrong? & \multirow{5}{*}{$1$} & \multirow{5}{*}{$2$}\\
    \textbf{Mental Counselor:} It seems that your daughter & & \\
    was not affected by what you told her. On the & & \\
    contrary, she felt that, well, she could comfort her & & \\
    mother. & & \\
  \bottomrule[1.2pt]
  \end{tabular}
  }
  \label{tab:freq}
\end{table}

Furthermore, the ER label quantifies the emotional reaction of the mental health counselor, 
whereas the CR label evaluates the cognitive reaction. Tables 6 and 7 present examples of 
misclassified instances in ER and CR labels respectively. Our observations indicate that the 
proposed model may produce inaccurate predictions when processing the counselor's long 
conversations composed of multiple short sentences. Due to the nature of their work, counselors 
often stop on important parts of the counseling process for extended periods. This will result in 
long dialogues with significant semantic spans, which poses challenges for the model's analysis. 
Therefore, future research will focus on enhancing our model's ability to predict empathy in 
the context of long conversations.

\section{Case Study}

\begin{table*}[ht]
\centering
\caption{An example showing training with and without supervisory documents assistance on EE labels}
\begin{tabularx}{\textwidth}{|X|p{\textwidth*3/4}|}
\hline
\textbf{Conversations} & \textbf{Client:} It seems like you are trying to make me feel less guilty.

\textbf{Mental Counselor:} Well, this is my true understanding. For example, if you are sitting in this consultation room, you can also understand that we are using each other. I am also using you to nourish me and increase my accumulation of cases. You can also use it, and you understand that you are using me to help you complete the task, whether it is to pass the level or to help you achieve self-growth. \\
\hline
\textbf{Supervisory Document of this Counseling Session} & The counselor once again made a strong confrontation and answered very straightforwardly about his understanding of the word ``use". His understanding of the two words was neutral, which meant it didn't matter whether it was good or bad. \textcolor{red}{\textbf{The client's response to this feedback was ``I feel much more relaxed". This relaxation also needs to be understood in terms of mental dynamics. This is a feeling that many people have. They feel that if they exploit others, it is equivalent to murdering others in their subconscious mind.} }The explanation for the origin of this feeling is that in my early years, in the relationship between my parents and me, the message conveyed by my parents was like this: child, if you continue to ask me to give you more things, I don't have them anymore. So if you want me to give you more things, it means murdering me. When parents have such a sense of scarcity when giving love or other things, the child's inner world will form another kind of scarcity, which is that after adulthood, they will not be able to make the most of the surrounding environment, including natural resources and interpersonal relationships, etc. This is why we see so many differences between people. Some people can use the environment to achieve their goals and make their lives better and better, while some people cannot. Some people can make the most of smartphones and use them as the best tool to help themselves, while some people only use part of the huge functions of mobile phones. Let’s continue watching the video. \\
\hline
\textbf{True EE Label} & \textbf{1 (weak expression)} \\
\hline
\textbf{Predicted EE Label with SDAT} & \textbf{1 (weak expression)} 
\\
\hline
\textbf{Predicted EE Label without SDAT} & \textbf{0 (no expression)} 
\\
\hline
\end{tabularx}
\label{tab:your_label}
\end{table*}

\begin{table*}[ht]
\centering
\caption{An example showing training with and without supervisory documents assistance on ER labels}
\begin{tabularx}{\textwidth}{|X|p{\textwidth*3/4}|}
\hline
\textbf{Conversations} & \textbf{Client:} Actually, I'm also very nervous. You said that my 
expression changed and my voice became different, but I didn’t feel it at all for I was very scared. 
It is just this feeling that you pushed me away. When you feel that these things have 
nothing to do with you, I feel a sense of being pushed away, a real sense of fear.

\textbf{Mental Counselor:} Emm.

\textbf{Client:} So do you want to express this? Just let me not count on you too much. You frowned, is it because I gave you a headache? Did I make you uncomfortable again?

\textbf{Mental Counselor:} Well, with this question of concern for me, our consultation is brought back to the beginning.\\
\hline
\textbf{Supervisory Document of this Counseling Session} & I watched this interview for a long time and finally found that the counselor's handling of the reality of this issue left a deep impression on me. The counselor said: ``With this question of concern for me, our consultation is brought back to the beginning." I think the reality of this issue has not been well responded to in this interview. I think everyone can respond more clearly. At this time, the counselor said that when you didn't get a definite answer from me, you would feel bad about yourself again, afraid of causing trouble for me or making me feel embarrassed. \textcolor{red}{\textbf{Although these questions sound like a reshaping of the questions raised by the client, we will find that the counselor here feels that he has not fully entered the client's heart to feel the state of this visit between the lines of his description. That is when there is no definite answer, such as this kind of uncertainty, and then this trouble, fear of causing trouble to others, fear of making the counselor feel embarrassed, this situation, and this emotion is not well experienced by the counselor.} } Then when the client has such emotions in his heart, I actually hope that the counselor can feel and respond to this emotion, and feel that the client's heart is already bitter enough. \\
\hline
\textbf{True ER Label} & \textbf{0 (no reaction)} \\
\hline
\textbf{Predicted ER Label with SDAT} & \textbf{0 (no reaction)} 
\\
\hline
\textbf{Predicted ER Label without SDAT} & \textbf{1 (weak reaction)} 
\\
\hline
\end{tabularx}
\label{tab:your_label}
\end{table*}

To further investigate the impact of supervisory documentation assisted training on the
experimental results, we randomly selected instances from the MEDIC dataset that were correctly 
classified with the assistance of supervised documentation for the case study. The results are shown 
in Tables 8, 9, and 10. Additionally, due to the lack of corresponding supervisory documents for the other 
Subreddits dataset, and the fact that it only contains textual modality information, we did not conduct 
ablation experiments or case studies on this dataset.

For the case with the EE label, as shown in Table 8, the background of this counseling session involves the 
client exploring feelings of guilt and self-worth in her intimate relationships. On the one hand, the client 
describes feelings of guilt toward her brother during her upbringing, as she received most of the family's love 
and affection. She perceives herself as not being good enough and unworthy of this love. On the other hand, she 
expresses pressure stemming from her husband's ``overly perfect" nature, feeling that she cannot balance this 
unequal relationship, as her husband bears most of the family responsibilities, making her contributions seem 
insignificant. The conversation in the table took place near the end of the session. After a lengthy 
consultation, the client expressed his feelings. In this 
segment of the conversation, the client experienced a sense of relaxation and a reduction in guilt, 
which was also emphasized in the corresponding supervisory documentation, as highlighted in red in 
the table. The supervisor provided a detailed psychological analysis of the emotions expressed by the 
client. Compared to networks trained without the assistance of supervisory documentation, our 
proposed method captured this additional information, leading to a correct classification for the EE 
label.

\begin{table*}[ht]
\centering
\caption{An example showing training with and without supervisory documents assistance on CR labels}
\begin{tabularx}{\textwidth}{|X|p{\textwidth*3/4}|}
\hline
\textbf{Conversations} & \textbf{Client:} So if this situation doesn't change, I think it may cause great harm to me and my child. 

\textbf{Mental Counselor:} Yes. On the one hand, you feel uncomfortable in this environment,  and on the other hand, you worry that your mother-in-law's raising may affect your child's future development.

\textbf{Client:} Yes.

\textbf{Mental Counselor:} So you said earlier that you hope your mother-in-law will leave your home.

\textbf{Client:} Exactly.

\textbf{Mental Counselor:} Hmm. Have you discussed this issue with your husband? How was the discussion? How did you express this to him?
\\
\hline
\textbf{Supervisory Document of this Counseling Session} & The client emphasized that if the current situation 
does not change if her mother-in-law does not change her parenting methods, it will cause great harm to the 
child's personality. I think this may be a little exaggerated. The client made such an exaggeration for two 
purposes. One goal is to exaggerate the cause of such harm so that she can continue to maintain the conflict 
with her mother-in-law. The second purpose is to put pressure on the counselor. We find that this client's way 
of dealing with conflict is to make the other party disappear, to make her mother-in-law leave her home. Of 
course, her desire to make the other party disappear if there is a conflict is also received by her husband, so 
her husband said, well if you let my mother leave, I will leave too. This is actually the result of the 
recognition of the client's strong inner desire for divorce. \textcolor{red}{\textbf{Our counselor grasped this 
point very well, and he continued to move in the direction of promoting the client's awareness and 
understanding.} } Let's continue watching the video. \\
\hline
\textbf{True CR Label} & \textbf{2 (strong expression)} \\
\hline
\textbf{Predicted CR Label with SDAT} & \textbf{2 (strong expression)} 
\\
\hline
\textbf{Predicted CR Label without SDAT} & \textbf{1 (weak expression)} 
\\
\hline
\end{tabularx}
\label{tab:your_label}
\end{table*}

In the example of the ER label, as shown in Table 9, the background of this counseling session is that the 
client shared her confusion regarding her intimate relationship with her boyfriend. She hoped to strengthen her 
connection with the counselor by increasing the frequency of sessions and seeking the counselor’s help in 
clarifying her concerns. During the discussion, the client expressed trust in the counselor and a desire for 
clear responses. However, the counselor, due to professional boundaries, declined her dependency, leading to 
the client’s anxiety and fear of causing inconvenience to the counselor. The conversation in the table occurred 
at the beginning of the session. It can be observed that the client exhibited significant emotional 
fluctuations at the beginning of the counseling session, characterized by intense nervousness and fear, as well 
as anxiety about disturbing the counselor. When confronted with the client's intense emotional fluctuations, the counselor in 
this conversation merely articulated some of their feelings, without adequately responding to the client's 
emotions or demonstrating a rich empathetic reaction. This shortcoming was also emphasized by the supervisor in 
the corresponding supervisory documentation, as highlighted in red in the table. The supervisor specifically 
pointed out that the counselor failed to fully engage with the client's inner world and lacked understanding 
and empathy toward the emotions expressed. Therefore, the true label for emotional reaction in this 
conversation is 0, indicating no emotional reaction. With the inclusion of this additional privileged 
information, we observed that our proposed method made more accurate predictions for the ER label compared to 
networks trained without the assistance of supervisory documentation.

In the example of the CR label, as shown in Table 10, the background of this counseling session is that the 
client perceives her mother-in-law to be overly involved in family life, while her husband consistently sides 
with his mother, leaving the client feeling isolated and helpless. The client also expresses concerns about her 
mother-in-law's parenting methods, believing they may have potential negative impacts on her child, which 
contributes to her confusion and distress within the marriage. The conversation in the table occurred at the 
very beginning of the counseling session, during which the client explained the reasons and background for 
seeking therapy. In response to the client's account, the counselor expressed a strong desire to further 
explore the issue. This was confirmed and explicitly endorsed by the supervisor in the supervision 
documentation, as highlighted in red and bold in the table. Therefore, the true label corresponding to the 
cognitive response for this conversation is 2, indicating a strong cognitive response. We found that with the 
inclusion of additional supervision information, our method was more accurate in predicting the intensity of 
empathy in the CR label compared to the network trained without the supervisory documentation assisted training.

Overall, the results of these case studies collectively demonstrate the importance of the supervisory 
documents assisted training. Whether in terms of overall accuracy comparisons or content-specific 
analyses of several test cases, our proposed method successfully extracts useful empathy-related information 
from the supervision documents, leading to more accurate empathy prediction results.

\section{Conclusion}
This paper presents an approach for predicting empathy levels in multi-modal scenarios with the 
assistance of supervisory documentation. Our method includes a multi-modal empathy prediction network 
for predicting empathy levels, as well as a supervisory documentation assisted training module.
The assisted module obtains the topic distribution of the supervisory documents, which contributes to 
constraining the text features for better prediction. Experimental results on 
two empathy datasets demonstrate the superiority of our proposed approach over existing works. 
We also conducted ablation studies to verify the effectiveness of each component. 
In future research, we plan to develop multi-task learning to improve the generalization capability 
of the framework.

\bibliographystyle{IEEEtran}
\bibliography{ref}

@inproceedings{mai2019divide,
  title={Divide, conquer and combine: Hierarchical feature fusion network with local and global perspectives for multimodal affective computing},
  author={Mai, Sijie and Hu, Haifeng and Xing, Songlong},
  booktitle={Proceedings of the 57th annual meeting of the association for computational linguistics},
  pages={481--492},
  year={2019}
}

@article{gibson2016deep,
  title={A deep learning approach to modeling empathy in addiction counseling},
  author={Gibson, James and Can, Dogan and Xiao, Bo and Imel, Zac E and Atkins, David C and Georgiou, Panayiotis and Narayanan, Shrikanth},
  journal={Commitment},
  volume={111},
  number={2016},
  pages={21},
  year={2016}
}

@misc{sharma2020computational,
      title={A Computational Approach to Understanding Empathy Expressed in Text-Based Mental Health Support}, 
      author={Ashish Sharma and Adam S. Miner and David C. Atkins and Tim Althoff},
      year={2020},
      eprint={2009.08441},
      archivePrefix={arXiv},
      primaryClass={cs.CL}
}

@inproceedings{xiao2012analyzing,
  title={Analyzing the language of therapist empathy in motivational interview based psychotherapy},
  author={Xiao, Bo and Can, Dogan and Georgiou, Panayiotis G and Atkins, David and Narayanan, Shrikanth S},
  booktitle={Proceedings of The 2012 Asia Pacific Signal and Information Processing Association Annual Summit and Conference},
  pages={1--4},
  year={2012},
  organization={IEEE}
}

@inproceedings{kumano2011analyzing,
  title={Analyzing empathetic interactions based on the probabilistic modeling of the co-occurrence patterns of facial expressions in group meetings},
  author={Kumano, Shiro and Otsuka, Kazuhiro and Mikami, Dan and Yamato, Junji},
  booktitle={2011 IEEE International Conference on Automatic Face \& Gesture Recognition (FG)},
  pages={43--50},
  year={2011},
  organization={IEEE}
}

@inproceedings{ter2020conversations,
  title={Conversations with documents: An exploration of document-centered assistance},
  author={ter Hoeve, Maartje and Sim, Robert and Nouri, Elnaz and Fourney, Adam and de Rijke, Maarten and White, Ryen W},
  booktitle={Proceedings of the 2020 Conference on Human Information Interaction and Retrieval},
  pages={43--52},
  year={2020}
}

@inproceedings{wang2007concept,
  title={Concept forest: A new ontology-assisted text document similarity measurement method},
  author={Wang, James Z and Taylor, William},
  booktitle={IEEE/WIC/ACM International Conference on Web Intelligence (WI'07)},
  pages={395--401},
  year={2007},
  organization={IEEE}
}

@inproceedings{li2011comment,
  title={Comment-guided learning: Bridging the knowledge gap between expert assessor and feature engineer},
  author={Li, Xiang and Lin, Wen-Pin and Ji, Heng},
  booktitle={Proc. International Conference on Advances in Information Mining and Management (IMMM2011)},
  year={2011}
}

@misc{guda2021empathbert,
      title={EmpathBERT: A BERT-based Framework for Demographic-aware Empathy Prediction}, 
      author={Bhanu Prakash Reddy Guda and Aparna Garimella and Niyati Chhaya},
      year={2021},
      eprint={2102.00272},
      archivePrefix={arXiv},
      primaryClass={cs.LG}
}

@article{devlin2018bert,
  title={Bert: Pre-training of deep bidirectional transformers for language understanding},
  author={Devlin, Jacob and Chang, Ming-Wei and Lee, Kenton and Toutanova, Kristina},
  journal={arXiv preprint arXiv:1810.04805},
  year={2018}
}

@article{vapnik2009new,
  title={A new learning paradigm: Learning using privileged information},
  author={Vapnik, Vladimir and Vashist, Akshay},
  journal={Neural networks},
  volume={22},
  number={5-6},
  pages={544--557},
  year={2009},
  publisher={Elsevier}
}

@inproceedings{vasava-etal-2022-transformer,
    title = "Transformer-based Architecture for Empathy Prediction and Emotion Classification",
    author = "Vasava, Himil  and
      Uikey, Pramegh  and
      Wasnik, Gaurav  and
      Sharma, Raksha",
    booktitle = "Proceedings of the 12th Workshop on Computational Approaches to Subjectivity, Sentiment {\&} Social Media Analysis",
    month = may,
    year = "2022",
    address = "Dublin, Ireland",
    publisher = "Association for Computational Linguistics",
    url = "https://aclanthology.org/2022.wassa-1.27",
    doi = "10.18653/v1/2022.wassa-1.27",
    pages = "261--264",
}

@inproceedings{dey2022enriching,
  title={Enriching deep learning with frame semantics for empathy classification in medical narrative essays},
  author={Dey, Priyanka and Girju, Roxana},
  booktitle={Proceedings of the 13th International Workshop on Health Text Mining and Information Analysis (LOUHI)},
  pages={207--217},
  year={2022}
}

@article{zhou2021language,
  title={The language of situational empathy},
  author={Zhou, Ke and Aiello, Luca Maria and Scepanovic, Sanja and Quercia, Daniele and Konrath, Sara},
  journal={Proceedings of the ACM on Human-Computer Interaction},
  volume={5},
  number={CSCW1},
  pages={1--19},
  year={2021},
  publisher={ACM New York, NY, USA}
}

@inproceedings{wu2021towards,
  title={Towards low-resource real-time assessment of empathy in counselling},
  author={Wu, Zixiu and Helaoui, Rim and Recupero, Diego Reforgiato and Riboni, Daniele},
  booktitle={Proceedings of the Seventh Workshop on Computational Linguistics and Clinical Psychology: Improving Access},
  pages={204--216},
  year={2021}
}

@article{montiel2022explainable,
  title={An Explainable Artificial Intelligence Approach for Detecting Empathy in Textual Communication},
  author={Montiel-V{\'a}zquez, Edwin Carlos and Ram{\'\i}rez Uresti, Jorge Adolfo and Loyola-Gonz{\'a}lez, Octavio},
  journal={Applied Sciences},
  volume={12},
  number={19},
  pages={9407},
  year={2022},
  publisher={MDPI}
}

@article{wu2022towards,
  title={Towards Automated Counselling Decision-Making: Remarks on Therapist Action Forecasting on the AnnoMI Dataset},
  author={Wu, Zixiu and Helaoui, Rim and Recupero, Diego Reforgiato and Riboni, Daniele},
  journal={Change},
  volume={25},
  pages={17},
  year={2022}
}

@inproceedings{lahnala2022caisa,
  title={CAISA at WASSA 2022: Adapter-Tuning for Empathy Prediction},
  author={Lahnala, Allison and Welch, Charles and Flek, Lucie},
  booktitle={Proceedings of the 12th Workshop on Computational Approaches to Subjectivity, Sentiment \& Social Media Analysis},
  pages={280--285},
  year={2022}
}

@inproceedings{vettigli2021empna,
  title={EmpNa at WASSA 2021: A Lightweight Model for the Prediction of Empathy, Distress and Emotions from Reactions to News Stories},
  author={Vettigli, Giuseppe and Sorgente, Antonio},
  booktitle={Proceedings of the Eleventh Workshop on Computational Approaches to Subjectivity, Sentiment and Social Media Analysis},
  pages={264--268},
  year={2021}
}

@inproceedings{shi2021modeling,
  title={Modeling clinical empathy in narrative essays},
  author={Shi, Shuju and Sun, Yinglun and Zavala, Jose and Moore, Jeffrey and Girju, Roxana},
  booktitle={2021 IEEE 15th International Conference on Semantic Computing (ICSC)},
  pages={215--220},
  year={2021},
  organization={IEEE}
}

@article{lee2022formality,
  title={Formality in psychotherapy: How are therapists’ and clients’ use of discourse particles related to therapist empathy?},
  author={Lee, Jonathan Him Nok and Chui, Harold and Lee, Tan and Luk, Sarah and Tao, Dehua and Lee, Nicolette Wing Tung},
  journal={Frontiers in Psychiatry},
  volume={13},
  year={2022},
  publisher={Frontiers Media SA}
}

@inproceedings{zhu2023medic,
  title={MEDIC: A multimodal empathy dataset in counseling},
  author={Zhu, Zhouan and Li, Chenguang and Pan, Jicai and Li, Xin and Xiao, Yufei and Chang, Yanan and Zheng, Feiyi and Wang, Shangfei},
  booktitle={Proceedings of the 31st ACM International Conference on Multimedia},
  pages={6054--6062},
  year={2023}
}

@inproceedings{mcfee2015librosa,
  title={librosa: Audio and music signal analysis in python},
  author={McFee, Brian and Raffel, Colin and Liang, Dawen and Ellis, Daniel P and McVicar, Matt and Battenberg, Eric and Nieto, Oriol},
  booktitle={Proceedings of the 14th python in science conference},
  volume={8},
  pages={18--25},
  year={2015}
}

@misc{cao2019openpose,
      title={OpenPose: Realtime Multi-Person 2D Pose Estimation using Part Affinity Fields}, 
      author={Zhe Cao and Gines Hidalgo and Tomas Simon and Shih-En Wei and Yaser Sheikh},
      year={2019},
      eprint={1812.08008},
      archivePrefix={arXiv},
      primaryClass={cs.CV}
}

@article{cui2021pre,
  title={Pre-training with whole word masking for chinese bert},
  author={Cui, Yiming and Che, Wanxiang and Liu, Ting and Qin, Bing and Yang, Ziqing},
  journal={IEEE/ACM Transactions on Audio, Speech, and Language Processing},
  volume={29},
  pages={3504--3514},
  year={2021},
  publisher={IEEE}
}

@article{cuff2016empathy,
  title={Empathy: A review of the concept},
  author={Cuff, Benjamin MP and Brown, Sarah J and Taylor, Laura and Howat, Douglas J},
  journal={Emotion review},
  volume={8},
  number={2},
  pages={144--153},
  year={2016},
  publisher={Sage Publications Sage UK: London, England}
}

@article{vaswani2017attention,
  title={Attention is all you need},
  author={Vaswani, A},
  journal={Advances in Neural Information Processing Systems},
  year={2017}
}

\end{document}